\documentclass[10pt,twocolumn,letterpaper]{article}

\usepackage{cvpr}
\usepackage{times}
\usepackage{epsfig}
\usepackage{graphicx}
\usepackage{amsmath}
\usepackage{amssymb}

\usepackage{stfloats}
\usepackage[x11names,dvipsnames,table]{xcolor} 
\usepackage{multirow}
\usepackage{array}
\newcolumntype{L}[1]{>{\raggedright\let\newline\\\arraybackslash\hspace{0pt}}m{#1}}
\newcolumntype{C}[1]{>{\centering\let\newline\\\arraybackslash\hspace{0pt}}m{#1}}
\newcolumntype{R}[1]{>{\raggedleft\let\newline\\\arraybackslash\hspace{0pt}}m{#1}}

\usepackage[breaklinks=true,bookmarks=false]{hyperref}

\cvprfinalcopy 

\def\cvprPaperID{****} 

\begin{document}

\title{Accelerating Deep Neural Networks with Spatial Bottleneck Modules}

\author{Junran Peng\textsuperscript{1}, Lingxi Xie\textsuperscript{2}, Zhaoxiang Zhang\textsuperscript{1}, Tieniu Tan\textsuperscript{1}, Jingdong Wang\textsuperscript{3}\\
\textsuperscript{1}Chinese Academy of Sciences\quad\textsuperscript{2}The Johns Hopkins University\quad\textsuperscript{3}Microsoft Research\\
\textsuperscript{1}{\tt\small \{pengjunran2015@,zhaoxiang.zhang@,tnt@nlpr.\}ia.ac.cn}\\
\textsuperscript{2}{\tt\small 198808xc@gmail.com}\quad\textsuperscript{3}{\tt\small jingdw@microsoft.com}
}

\maketitle

\begin{abstract}
This paper presents an efficient module named {\bf spatial bottleneck} for accelerating the convolutional layers in deep neural networks. The core idea is to decompose convolution into two stages, which first reduce the spatial resolution of the feature map, and then restore it to the desired size. This operation decreases the sampling density in the spatial domain, which is independent yet complementary to network acceleration approaches in the channel domain. Using different sampling rates, we can tradeoff between recognition accuracy and model complexity.\\
As a basic building block, spatial bottleneck can be used to replace any single convolutional layer, or the combination of two convolutional layers. We empirically verify the effectiveness of spatial bottleneck by applying it to the deep residual networks. Spatial bottleneck achieves $2\times$ and $1.4\times$ speedup on the regular and channel-bottlenecked residual blocks, respectively, with the accuracies retained in recognizing low-resolution images, and even improved in recognizing high-resolution images.
\end{abstract}

\section{Introduction}
\label{Introduction}

In the modern era, with the availability of large-scale datasets~\cite{deng2009imagenet}\cite{lin2014microsoft} and powerful computational resources, deep learning techniques especially the convolutional neural networks (CNNs) have been applied to a wide range of computer vision tasks, such as image classification~\cite{krizhevsky2012imagenet}, object detection~\cite{girshick2014rich}, semantic segmentation~\cite{long2015fully}, edge detection~\cite{xie2015holistically}, {\em etc}. Despite their great success, we still care about an important issue, which is the expensive computational overhead which limits them from being applied in some real-time scenarios.

Accelerating convolutional neural networks has been attracting a lot of interests. Convolution is the most computationally expensive module in each network, and its FLOPs (the number of floating point operations) is proportional to the size of convolutional kernels, the number of input and output channels, and the spatial resolution of the output feature map. A lot of efforts have been made to eliminate the {\em redundancy} in the computation. There are two main research lines, namely, compressing pre-trained networks and designing more efficient structures. Most existing approaches work on the filter level, {\em i.e.}, accelerating convolution by reducing the number and/or precision of the multiplication and addition operations for each output. Among these, compression algorithms include sparsifying convolutional kernels~\cite{liu2015sparse}, pruning filters or channels~\cite{he2017channel}, learning low-precision filter weights~\cite{rastegari2016xnor}, {\em etc.}, and structure design techniques include Xception~\cite{chollet2017xception}, interleaved group convolutions~\cite{zhang2017interleaved}, bottlenecked convolution~\cite{he2016deep}, {\em etc}. Despite these methods, we note that reducing the factor of spatial resolution, while being a promising direction, is rarely studied before.

This paper aims at reducing computational costs in the spatial domain. This is independent yet complementary to the efforts made in the channel domain. We propose a simple building block named {\bf spatial bottleneck}, which consists of a convolutional layer which down-samples the input feature map to a smaller size ({\em e.g.}, half width and height), and a deconvolutional layer which up-samples the feature map back to the desired size (most often, the original size). In this way, we share a part of computation, and sparsify the sampling rate in the spatial domain. By controlling the {\em density} of spatial sampling, we can achieve different tradeoffs between recognition accuracy and model complexity.

Spatial bottleneck is a generalized module which can be used to replace any single convolutional layer or the combination of two convolutional layers\footnote{When spatial bottleneck is used to replace one convolutional layer, we do not insert non-linear normalization and activation operations between convolution and deconvolution. This guarantees the linearity of spatial bottleneck, and thus the network depth is not increased. When two convolutional layers are replaced at once, we preserve all the non-linear functions between them.}. We provide an example based on the deep residual networks~\cite{he2016deep}. Each residual block, either with or without channel bottleneck\footnote{We refer to the bottleneck module in~\cite{he2016deep} by {\em channel} bottleneck in order to discriminate it from the proposed {\em spatial} bottleneck module.}, can be modified into a spatial bottleneck block. Each spatial bottleneck blocks enjoy a reduced FLOPs and, consequently, a favorable speed ($2\times$ speedup on a regular residual block, or $1.4\times$ speedup on a channel-bottlenecked residual block). We empirically evaluate our approach on both the CIFAR~\cite{krizhevsky2009learning} and ILSVRC2012~\cite{russakovsky2015imagenet} datasets, and show that the accelerated networks retain classification accuracy in low-resolution datasets, and perform better in recognizing high-resolution images.

\section{Related Work}
\label{RelatedWork}

\subsection{Deep Convolutional Neural Networks}
\label{RelatedWork:DeepLearning}

The research field of computer vision, in particular image classification, has been dominated by deep convolutional neural networks. These models mostly rely on powerful computational resources such as modern GPUs, and large-scale datasets such as ImageNet~\cite{deng2009imagenet} or MS-COCO~\cite{lin2014microsoft}. The fundamental principle is to build a hierarchical structure to learn and represent image features. In the early years, these deep networks were first applied to simple recognition tasks~\cite{lecun1998gradient}\cite{krizhevsky2009learning}, but recently, with the development of activation~\cite{nair2010rectified} and regularization~\cite{srivastava2014dropout} techniques, researchers were able to train deeper network architectures~\cite{krizhevsky2009learning}\cite{simonyan2015very}\cite{szegedy2015going} towards human-level recognition performance on high-resolution natural images. It is believed that deeper networks can produce better recognition performance~\cite{he2016deep}\cite{szegedy2016rethinking}\cite{wang2016deeply}\cite{zhao2016deep}\cite{chen2017dual}\cite{hu2017squeeze}\cite{huang2017densely}\cite{xie2017aggregated}. Training very deep ({\em e.g.}, more than $100$ layers) networks requires an approach named batch normalization~\cite{ioffe2015batch} to avoid the neural responses from going out of control. Beyond manually designing neural networks, researchers also explored the possibility of learning network architectures from training data automatically~\cite{xie2017genetic}\cite{zoph2017neural}.

The fundamental advances in image classification helped other vision tasks. It was verified that visual features extracted from deep networks are more effective than those generated by conventional approaches~\cite{donahue2014decaf}\cite{razavian2014cnn}. In addition, the pre-trained networks in image classification can be transferred to other problems via fine-tuning, including fine-grained classification~\cite{lin2015bilinear}, object detection~\cite{girshick2014rich}\cite{girshick2015fast}\cite{ren2015faster}, semantic segmentation~\cite{long2015fully}\cite{chen2016deeplab}, edge detection~\cite{xie2015holistically}, pose estimation~\cite{toshev2014deeppose}, {\em etc}.

\subsection{Accelerating Deep Neural Networks}
\label{RelatedWork:Acceleration}

Despite the great success, deep neural networks suffer heavy computational overheads, which limit them from being deployed to real-time applications or on mobile devices. People mostly care about the convolutional layers which account for most of computational overheads. Some basic principles have been proposed for designing the overall architecture, {\em e.g.}, decreasing the spatial resolution as increasing the number of channels~\cite{krizhevsky2012imagenet}\cite{simonyan2015very}, and using a small number of convolutional layers at the low levels to down-sample the feature map rapidly~\cite{simonyan2015very}\cite{he2016deep}.
There are also efforts in accelerating each convolutional layer individually. Note that the FLOPs of convolution is proportional to (i) the kernel size, (ii) the numbers of connected input-output channel pairs\footnote{In a normal implementation, this number equals to the product of the numbers of input and output channels.}, and (iii) the spatial resolution of the output feature map.

Regarding the first point, most modern deep networks use small ($3\times3$) convolutional kernels, except for in the scenarios of constructing auxiliary connections~\cite{szegedy2015going} or shrinking the feature map size at the beginning of the network~\cite{krizhevsky2012imagenet}\cite{he2016deep}. Larger kernels are considered more expensive and rarely used.

The second choice involves reducing the number of output channels, or sparsifying the input-output connections in the channel domain. The former type includes the channel-bottlenecked blocks used in very deep residual networks~\cite{he2016deep}\cite{hu2017squeeze}, and the effort in pruning redundant channels after the network is trained~\cite{he2017channel}. The latter type includes the use of group convolution~\cite{chollet2017xception}\cite{xie2017aggregated}\cite{zhang2017interleaved} and/or sparse convolution~\cite{liu2015sparse}.

However, we see fewer studies in exploring the third possibility, {\em i.e.}, reducing the spatial resolution of the output layer. There were some efforts but none of them were focused on acceleration. Related work includes the encoder-decoder networks~\cite{hinton2006reducing} which first down-sampled the feature map to recognize the object, and then up-sampled it to allow fine-scaled detection~\cite{newell2016stacked} or segmentation~\cite{long2015fully}\cite{chen2016deeplab}. Our idea, spatial bottleneck, is composed of one convolutional (down-sampling or encoding) layer and one deconvolutional (up-sampling or decoding) layer. It is a standalone block, which can be applied to any convolutional layer and reduce the FLOPs of convolution by a half.

\section{Our Approach}
\label{Approach}

\begin{figure*}[t]
\begin{center}
    \includegraphics[width=14.0cm]{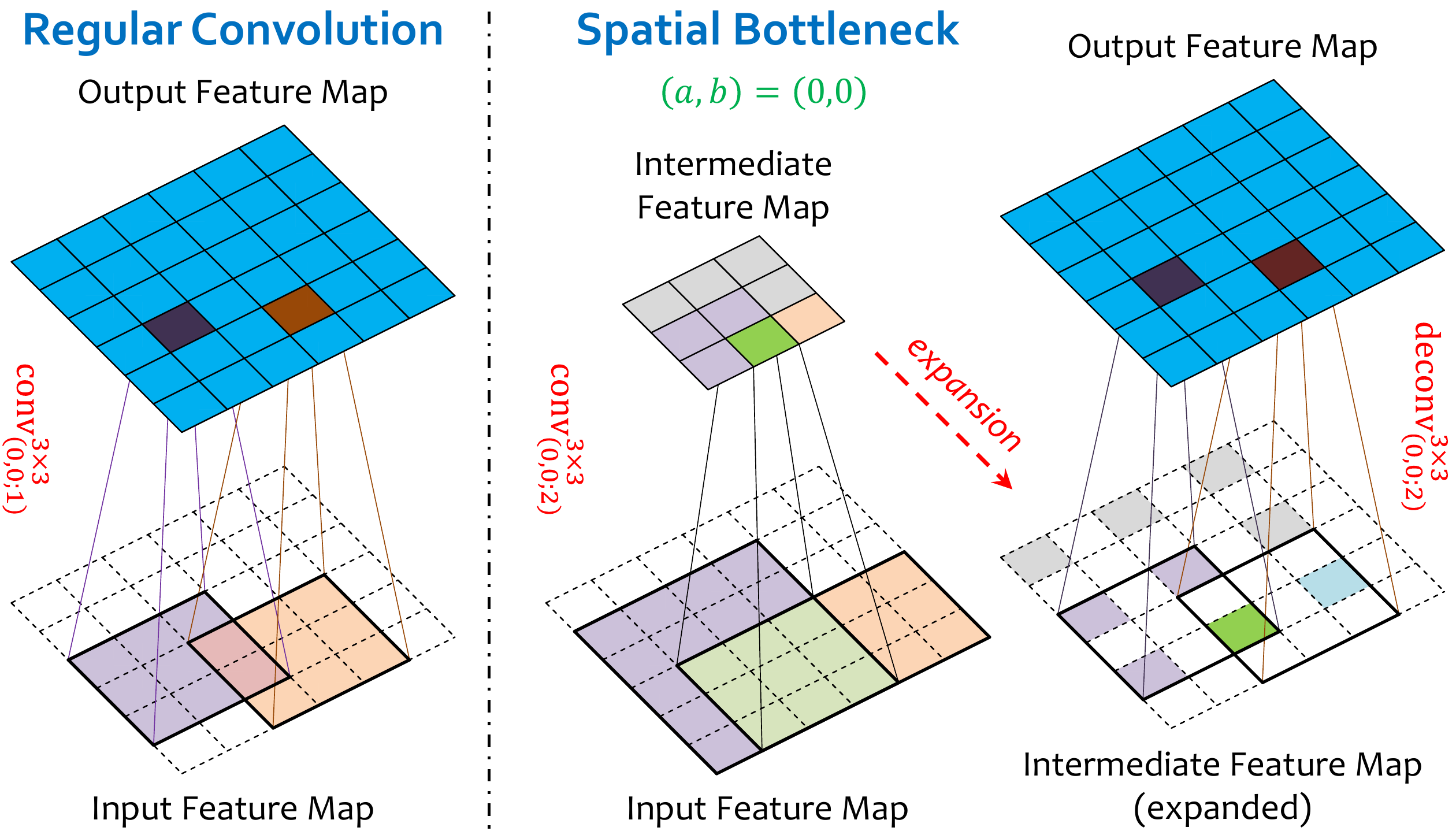}
\end{center}
\caption{
    Illustration of a regular convolutional layer $\mathrm{conv}_{\left(0,0;1\right)}^{3\times3}$ (left) and a spatial bottleneck module $\mathrm{SB}_{\left(0,0;2\right)}^{3\times3}$ (right). $\mathrm{SB}_{\left(0,0;2\right)}^{3\times3}$ is decomposed into $\mathrm{conv}_{\left(0,0;2\right)}^{3\times3}$ and $\mathrm{deconv}_{\left(0,0;2\right)}^{3\times3}$. The receptive field sizes of the two output neurons are increased from $3\times3$ to $5\times4$ and $3\times5$, respectively, with part of computation (the magenta region on the input map) being shared. This figure is best viewed in color.
}
\label{Fig:Module}
\end{figure*}

\subsection{The Spatial Bottleneck Module}
\label{Approach:Module}

In a convolutional layer, the input feature map $\mathbf{X}$ is a $W_1\times H_1\times D_1$ cube, with $W_1$, $H_1$ and $D_1$ indicating its width, height and depth (also referred to as the number of channels), respectively. The output feature map, similarly, is a cube $\mathbf{Z}$ with $W_2\times H_2\times D_2$ entries. The convolution ${\mathbf{Z}}={\mathbf{f}\!\left(\mathbf{X}\right)}$ is parameterized by $D_2$ convolutional kernels, each of which is a $S\times S\times D_1$ cube. The number of floating-point operations (FLOPs, each multiplication followed by a summation is counted by $1$) is $W_2H_2D_1D_2S^2$. There are three factors, namely, the kernel size ($S^2$), the number of connections in the channel domain ($D_1D_2$), and the resolution of the output feature map ($W_2H_2$). The proposed {\bf spatial bottleneck} module is aimed at decreasing the factor of $W_2H_2$ -- it is independent of, and thus complementary to the methods to optimize other two factors, {\em e.g.}, using group convolution to reduce the factor of $D_1D_2$~\cite{xie2017aggregated}.

The core idea of decreasing $W_2H_2$ is to first reduce the spatial resolution of the feature map, and then restore it to the desired size. In practice, we implement ${\mathbf{Y}}={\mathbf{f}_1\!\left(\mathbf{X}\right)}$ as a stride-$K$ convolution, and ${\mathbf{Z}}={\mathbf{f}_2\!\left(\mathbf{Y}\right)}$ as a stride-$K$ deconvolution. The width and height of $\mathbf{Y}$ are $1/K$ of $\mathbf{X}$ and $\mathbf{Z}$, and so both of these operations need $1/K^2$ computational costs of the original convolution. We denote these operations by $\mathrm{conv}_{\left(a,b;K\right)}^{S\times S}$ and $\mathrm{deconv}_{\left(a,b;K\right)}^{S\times S}$, respectively, where $\left(a,b\right)$ is a pair of integers satisfying ${0}\leqslant{a,b}<K$ and indicating the starting index of convolution. The entire spatial bottleneck module, denoted by $\mathrm{SB}_{\left(a,b;K\right)}^{S\times S}$, requires roughly $2W_2H_2D_1D_2S^2/K^2$ FLOPs\footnote{This number may be slightly different when $W_2$ or $H_2$ is not divisible by $K$.}. Note that, although the original convolution ${\mathbf{Z}}={\mathbf{f}\!\left(\mathbf{X}\right)}$ is decomposed into two layers, namely ${\mathbf{Z}}={\mathbf{f}_2\circ\mathbf{f}_1\!\left(\mathbf{X}\right)}$, as we do not insert non-linearity between $\mathbf{f}_1\!\left(\cdot\right)$ and $\mathbf{f}_2\!\left(\cdot\right)$ when it is used to replace a normal convolution layer, spatial bottleneck is still a linear operation and thus the network depth remains unchanged.

Another way of understanding spatial bottleneck is to assume the intermediate layer $\mathbf{Y}'$ has the same spatial resolution as $\mathbf{X}$ and $\mathbf{Z}$, but only a subset of spatial positions on $\mathbf{Y}'$ are sampled, {\em i.e.}, convolution is computed at all coordinates $\left(w,h\right)$ satisfying ${\left(w,h\right)}\equiv{\left(a,b\right)}\,{\left(\mathrm{mod}\,K\right)}$. We will discuss on the sampling strategy in the following sections.

As shown in Figure~\ref{Fig:Module}, spatial bottleneck allows part of computations to be shared by some neurons in the output feature map. This strategy enlarges the receptive field of neurons with reduced computational costs.

\subsection{Integrating Spatial Bottleneck into Residual Blocks}
\label{Approach:Applications}

\begin{figure*}[t]
\begin{center}
    \includegraphics[width=14.0cm]{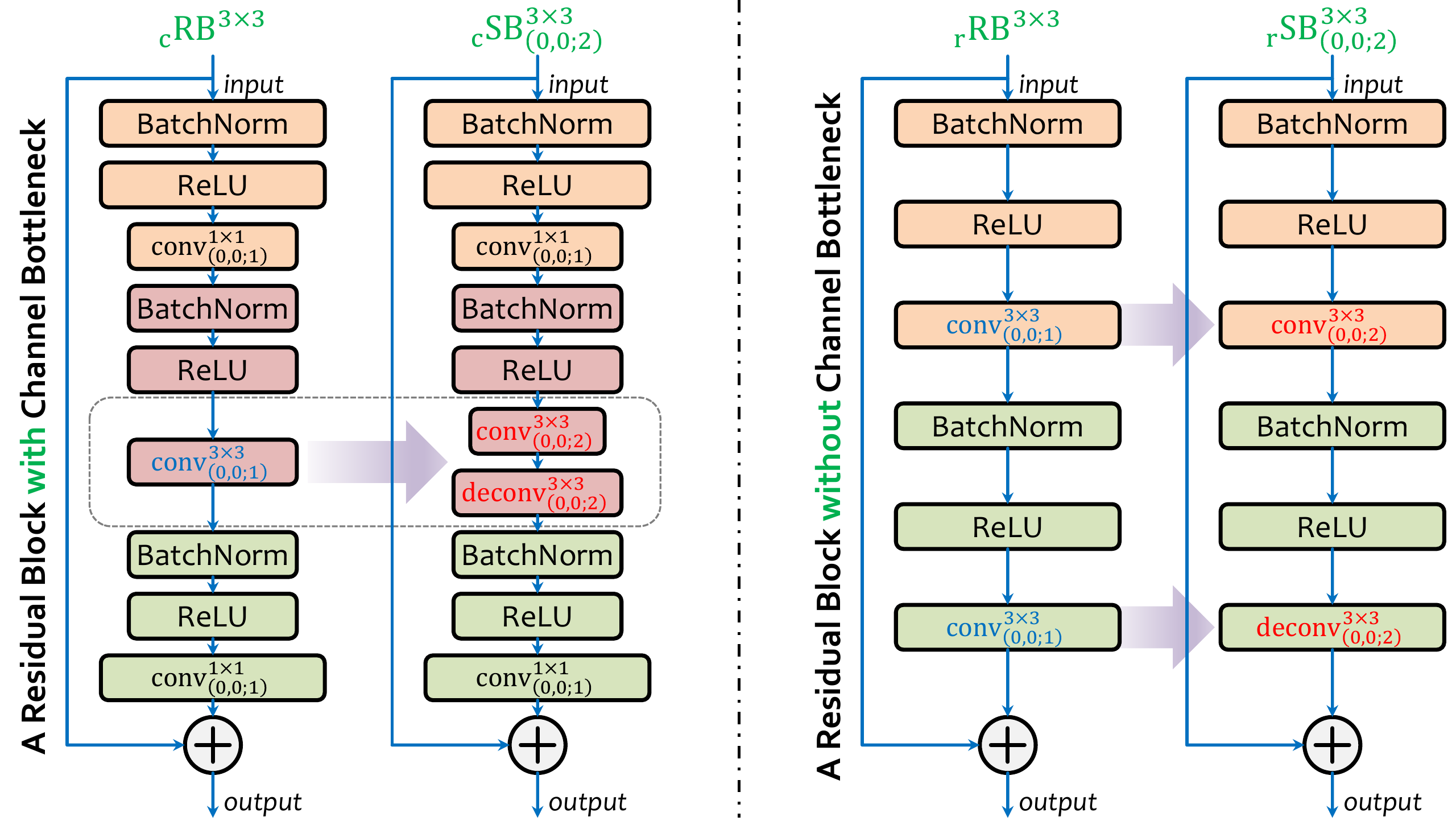}
\end{center}
\caption{
    Modifying a residual block with (left) or without (right) a channel bottleneck into a spatial bottleneck module (best viewed in color). The width of each rectangle indicates the number of channels in the corresponding layer.
}
\label{Fig:Applications}
\end{figure*}

As a practical example, we apply spatial bottleneck to deep residual networks. It can also be applied to other networks with residual blocks, such as~\cite{xie2017aggregated}\cite{han2017deep}\cite{hu2017squeeze}\cite{huang2017densely}.

We start with the case that channel bottleneck is present (see the left part of Figure~\ref{Fig:Applications}), in which a $1\times1$ convolution is first applied to reduce the number of channels to $1/C$, and then a regular $3\times3$ convolution works on the channel-reduced layer, followed by another $1\times1$ convolution to restore the desired number of channels. This residual block is denoted by $_\mathrm{c}\mathrm{RB}^{3\times3}$. We use a spatial bottleneck module to replace $\mathrm{conv}_{\left(0,0;1\right)}^{3\times3}$ with $\mathrm{SB}_{\left(0,0;2\right)}^{3\times3}$. In order not to increase the number of parameters, we reduce the channel number of the first convolution in spatial bottleneck by half. Spatial bottleneck reduces the FLOPs of $\mathrm{conv}_{\left(0,0;1\right)}^{3\times3}$ by $75\%$, and that of the entire residual block by $27/\left(8C+36\right)$. We denote this modified block by $_\mathrm{c}\mathrm{SB}_{\left(a,b;2\right)}^{3\times3}$.

In another case when channel bottleneck is not present (see the right part of Figure~\ref{Fig:Applications}), there are two regular $3\times3$ convolutional layers in a residual block. This residual block is denoted by $_\mathrm{r}\mathrm{RB}^{3\times3}$. A natural choice is to replace each of them with a spatial bottleneck module, but we find that it is a more efficient alternative to use one spatial bottleneck in the residual block, {\em i.e.}, using the stride-$2$ convolutional and deconvolutional layers to replace the two original convolutional layers, respectively. Spatial bottleneck reduces the FLOPs of the entire block by $75\%$. We denote this modified block by $_\mathrm{r}\mathrm{SB}_{\left(a,b;2\right)}^{3\times3}$.

\subsection{Chessboard Sampling and Receptive Fields}
\label{Approach:Sampling}

\begin{figure}[t]
\begin{center}
    \includegraphics[width=8.0cm]{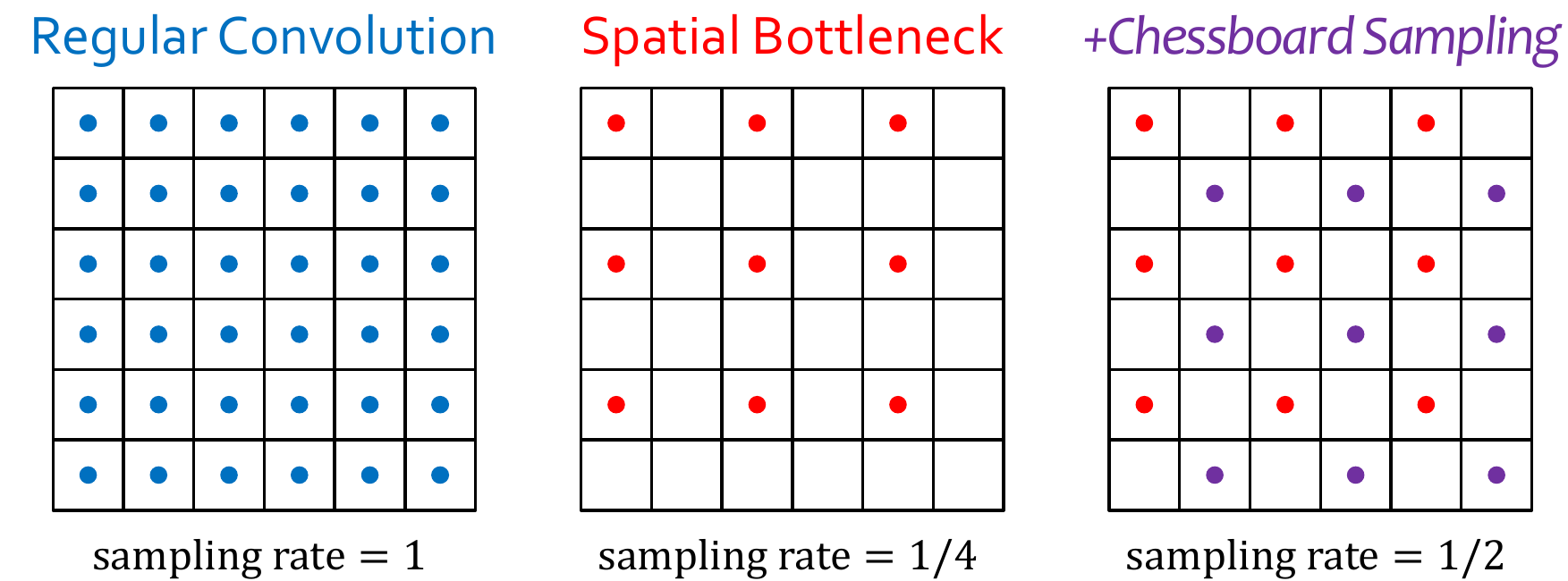}
\end{center}
\caption{
    Illustration of different sampling strategies. A black dot indicates that the corresponding position is sampled. In this example we show a $6\times6$ feature map, but our approach can fit an arbitrary size in practice.
}
\label{Fig:Sampling}
\end{figure}

By setting ${K}={2}$, spatial bottleneck only samples $1/4$ of spatial positions. This harms classification accuracy due to information loss. An intuitive approach is to sample more spatial positions. For example, in a regular residual block $_\mathrm{r}\mathrm{SB}^{3\times3}$, we sample all positions satisfying ${\left(w,h\right)}\equiv{\left(0,0\right)}\,{\left(\mathrm{mod}\,2\right)}$ or ${\left(w,h\right)}\equiv{\left(1,1\right)}\,{\left(\mathrm{mod}\,2\right)}$. This strategy, named {\em chessboard sampling}, is denoted by $_\mathrm{r}\mathrm{SB}_{\left(0,0;2\right)\vee\left(1,1;2\right)}^{3\times3}$. It increases the sampling rate from $1/4$ to $1/2$, bringing accuracy gain at the price of doubled computational costs compared to $_\mathrm{r}\mathrm{SB}_{\left(0,0;2\right)}^{3\times3}$. Different sampling strategies are illustrated in Figure~\ref{Fig:Sampling}. We can continue increasing the sampling rate, but it helps little to classification because the {\em saturation} of spatial information (see experimental results in Table~\ref{Tab:ResultsCIFARDensity}).

We provide another perspective from the receptive field. First note that all output neurons in a regular convolutional layer have the same receptive field of $3\times3$ in the input feature map. However, when spatial bottleneck is applied, the receptive field of each neuron can vary from $3\times3$ to $5\times5$. Such variance is {\em not} friendly to network training~\cite{xie2016geometric}. After chessboard sampling is used, all neurons have the same receptive field size ($5\times5$) again. By sharing computation, spatial bottleneck enlarges the receptive fields using reduced costs.

\subsection{Relationship to Previous Work}
\label{Approach:Relationship}

We discuss the relationship between our approach and a series of previous work that changes the spatial properties of feature maps for various purposes.

Our approach is related to a series of approaches based on the encoder-decoder networks. Typical examples include the auto-encoder~\cite{hinton2006reducing}, the fully-convolutional networks~\cite{long2015fully}, the U-net~\cite{ronneberger2015u}, the stacked hourglass networks~\cite{newell2016stacked}, {\em etc}. These networks first use several down-sampling layers for encoding (recognizing the object), and then up-sample the result for decoding (describing the object at a finer scale, {\em e.g.}, segmentation or pose estimation). Spatial bottleneck is in some sense similar, and first reduces the spatial resolution by convolution and then restores it by deconvolution. There are some differences. (i) The purpose of spatial bottleneck is different and to accelerate the computation, which is rarely studied before; and (ii) our approach uses de-convolution for up-sampling rather than bilinear interpolation~\cite{newell2016stacked} or up-convolution~\cite{ronneberger2015u}. In particular, the spatial bottleneck module is a drop-in replacement of one convolution or a pair of convolutions, and can be adopted in existing CNNs for acceleration.

Reducing the resolution for adjusting computation cost has essentially widely adopted in the current standard networks, including VGGNet, GoogleNet, and ResNet, to form multi-stage networks, where the resolutions are decreased stage by stage. Differently, our approach performs resolution reduction by convolution and immediate resolution increasing by deconvolution. Our approach can be directly combined into those networks, {\em e.g.}, applied to ResNets, which is used as an example application in this paper.



\section{CIFAR Experiments}
\label{ExperimentsCIFAR}

\newcommand{\colwidthA}{1.7cm}
\newcommand{\colwidthB}{1.9cm}
\newcommand{\colwidthC}{1.3cm}
\begin{table*}[!btp]
\centering
\begin{tabular}{|l||R{\colwidthA}|R{\colwidthA}||R{\colwidthB}|R{\colwidthB}||R{\colwidthC}|R{\colwidthC}|}
\hline
      &                             \multicolumn{2}{c||}{CIFAR10} &                            \multicolumn{2}{c||}{CIFAR100} &                                \multicolumn{2}{c|}{FLOPs} \\
\hline
Depth &  ${}_\mathrm{r}\mathrm{RN}$ & ${}_\mathrm{r}\mathrm{SBN}$ &  ${}_\mathrm{r}\mathrm{RN}$ & ${}_\mathrm{r}\mathrm{SBN}$ &  ${}_\mathrm{r}\mathrm{RN}$ & ${}_\mathrm{r}\mathrm{SBN}$ \\
\hline\hline
$20$  &     $\mathbf{ 6.99}\pm0.18$ &              $ 7.15\pm0.22$ &     $\mathbf{30.10}\pm0.57$ &              $30.57\pm0.23$ &           $ 40.5\mathrm{M}$ &  $\mathbf{ 24.0}\mathrm{M}$ \\
\hline
$44$  &              $ 5.86\pm0.28$ &     $\mathbf{ 5.80}\pm0.10$ &     $\mathbf{27.28}\pm0.33$ &              $27.35\pm0.07$ &           $ 97.2\mathrm{M}$ &  $\mathbf{ 52.3}\mathrm{M}$ \\
\hline
$62$  &              $ 5.70\pm0.11$ &     $\mathbf{ 5.50}\pm0.13$ &              $26.28\pm0.32$ &     $\mathbf{26.27}\pm0.28$ &           $139.6\mathrm{M}$ &  $\mathbf{ 73.6}\mathrm{M}$ \\
\hline
$86$  &              $ 5.32\pm0.11$ &     $\mathbf{ 5.31}\pm0.17$ &              $25.48\pm0.28$ &     $\mathbf{25.25}\pm0.31$ &           $193.6\mathrm{M}$ &  $\mathbf{101.9}\mathrm{M}$ \\
\hline
$110$ &              $ 5.13\pm0.18$ &     $\mathbf{ 5.12}\pm0.14$ &     $\mathbf{25.01}\pm0.32$ &              $25.13\pm0.01$ &           $252.9\mathrm{M}$ &  $\mathbf{130.0}\mathrm{M}$ \\
\hline\hline\hline
      &                             \multicolumn{2}{c||}{CIFAR10} &                            \multicolumn{2}{c||}{CIFAR100} &                                \multicolumn{2}{c|}{FLOPs} \\
\hline
Depth &  ${}_\mathrm{c}\mathrm{RN}$ & ${}_\mathrm{c}\mathrm{SBN}$ &  ${}_\mathrm{c}\mathrm{RN}$ & ${}_\mathrm{c}\mathrm{SBN}$ &  ${}_\mathrm{c}\mathrm{RN}$ & ${}_\mathrm{c}\mathrm{SBN}$ \\
\hline\hline
$29$  &              $ 6.42\pm0.11$ &     $\mathbf{ 6.27}\pm0.11$ &              $27.04\pm0.17$ &     $\mathbf{26.35}\pm0.12$ &           $ 34.5\mathrm{M}$ &  $\mathbf{ 27.5}\mathrm{M}$ \\
\hline
$47$  &              $ 5.56\pm0.26$ &     $\mathbf{ 5.51}\pm0.09$ &              $24.68\pm0.20$ &     $\mathbf{24.67}\pm0.21$ &           $ 57.1\mathrm{M}$ &  $\mathbf{ 42.9}\mathrm{M}$ \\
\hline
$65$  &              $ 5.15\pm0.10$ &     $\mathbf{ 5.09}\pm0.07$ &              $24.04\pm0.18$ &     $\mathbf{23.98}\pm0.12$ &           $ 79.6\mathrm{M}$ &  $\mathbf{ 58.4}\mathrm{M}$ \\
\hline
$83$  &     $\mathbf{ 4.84}\pm0.11$ &              $ 4.96\pm0.04$ &     $\mathbf{23.17}\pm0.28$ &              $23.26\pm0.16$ &           $102.3\mathrm{M}$ &  $\mathbf{ 73.8}\mathrm{M}$ \\
\hline
$101$ &     $\mathbf{ 4.76}\pm0.04$ &              $ 5.03\pm0.12$ &     $\mathbf{22.96}\pm0.02$ &              $23.16\pm0.04$ &           $125.0\mathrm{M}$ &  $\mathbf{ 89.3}\mathrm{M}$ \\
\hline
\end{tabular}
\caption{
    Classification error rate ($\%$) and FLOPs of different networks. All the error rates reported are the average over $5$ individual runs. Each architecture has almost the same FLOPs for CIFAR10 and CIFAR100.
}
\label{Tab:ResultsCIFAR}
\end{table*}

\subsection{Settings}
\label{ExperimentsCIFAR:Settings}

The CIFAR datasets~\cite{krizhevsky2009learning} were sampled from a large-scale small picture dataset. There are two subsets, known as CIFAR10 and CIFAR100. Each set has $60\rm{,}000$ low-resolution ($32\times32$) RGB images, in which $50\rm{,}000$ are used for training and $10\rm{,}000$ for testing. Both the training and testing samples are evenly distributed over all $10$ or $100$ classes.

We use the deep residual networks~\cite{he2016deep} with pre-activations~\cite{he2016identity} as our baseline. Based on the residual blocks with or without channel bottleneck, we construct $5$ variants with different depths by stacking different numbers of residual blocks. The spatial bottleneck network is constructed by replacing all residual blocks with the corresponding spatial bottleneck blocks shown in Figure~\ref{Fig:Applications}. Spatial bottleneck reduces the FLOPs by around $27\%$ and $50\%$ for each residual block with and without channel bottlenecks, respectively.

We train all these networks from scratch. Standard Stochastic Gradient Descent (SGD) with a Nesterov momentum of $0.9$ is used. Each network is trained for $400$ epochs. The initial learning rate is set to be $0.1$, and divided by $10$ after $200$ and $300$ epochs. The size of each mini-batch is $64$, and the weight decay is $0.0001$. Standard data augmentations are performed in training. A $4$-pixel margin is added to all four sides of each image, which is enlarged from $32\times32$ to $40\times40$. The pixels within the margin are filled up by the symmetric pixels in the original image. Then, we crop a $32\times32$ image and flip it with a probability of $0.5$. No augmentation is used in the testing process.

\subsection{Results}
\label{ExperimentsCIFAR:Results}

\begin{figure*}[t]
\begin{center}
    \includegraphics[width=5.5cm]{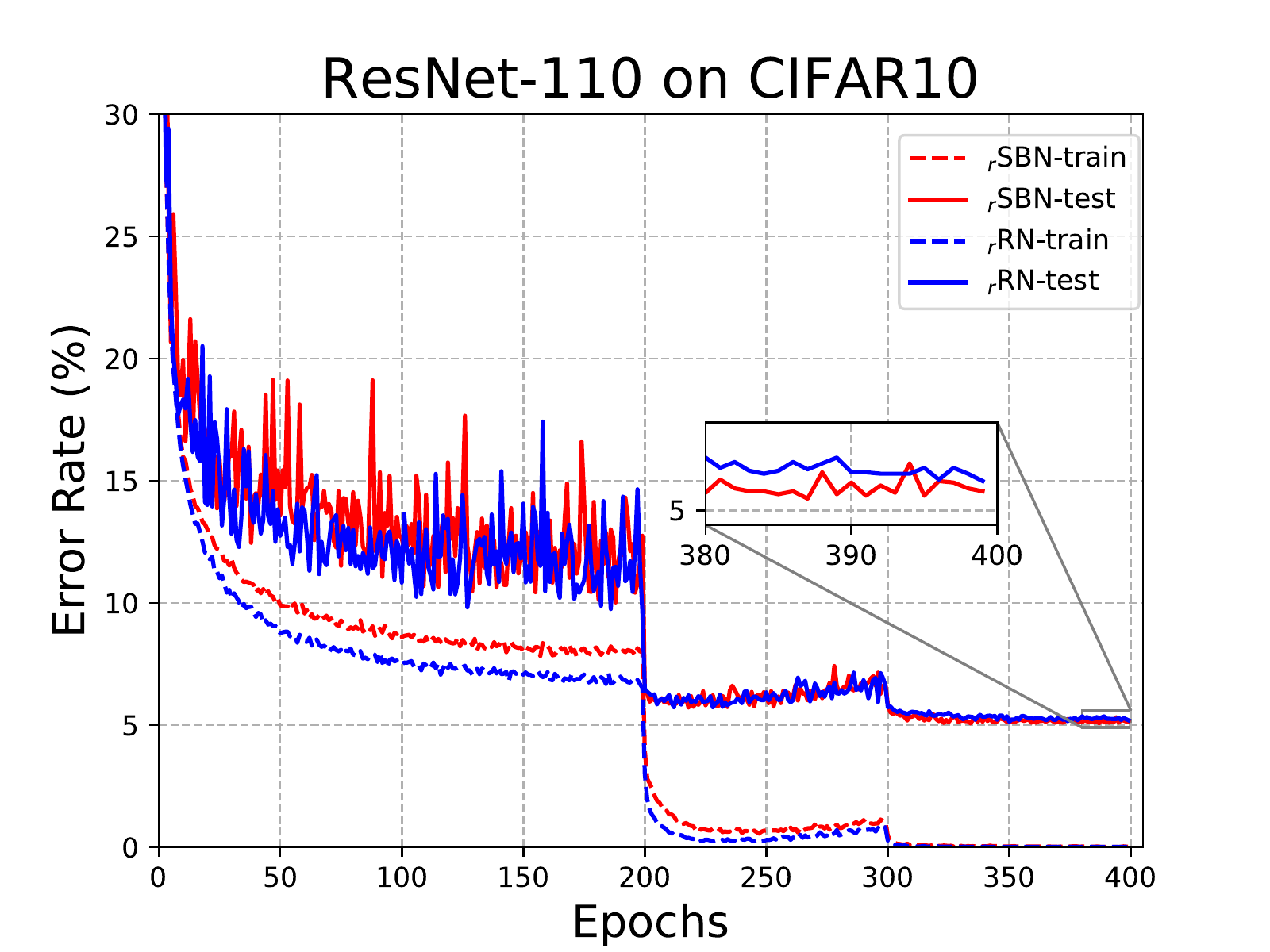}
    \includegraphics[width=5.5cm]{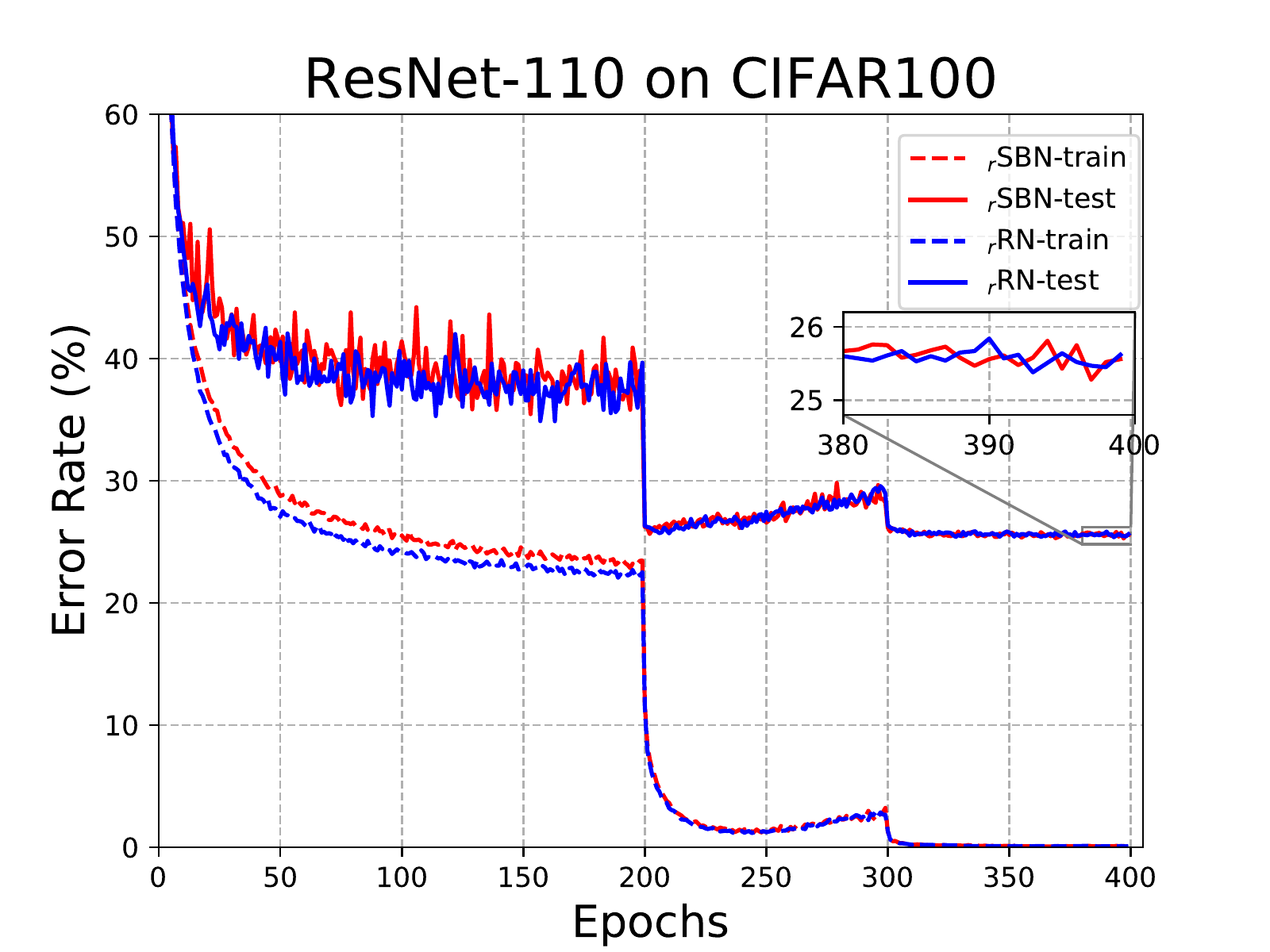}
    \includegraphics[width=5.5cm]{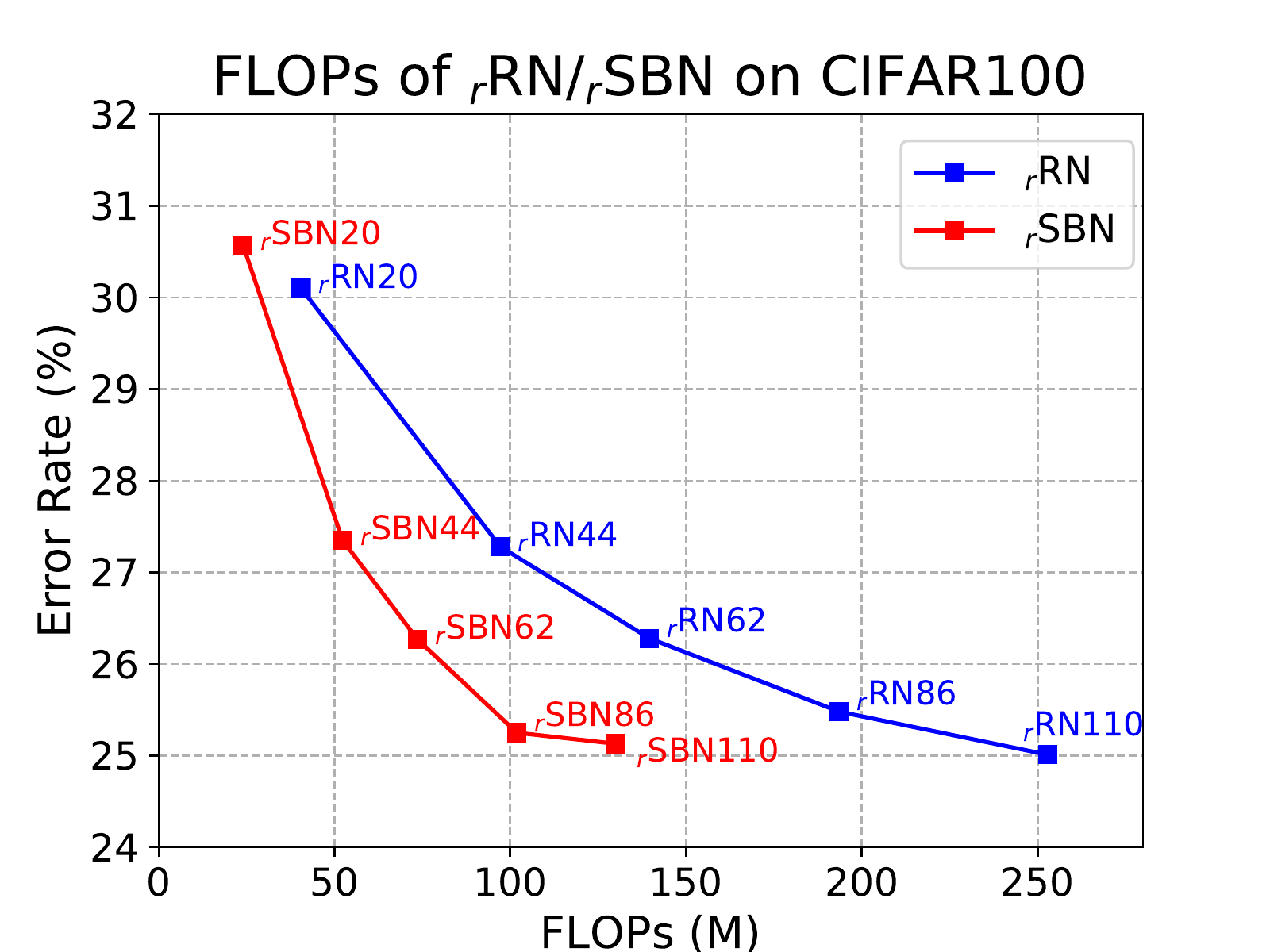}\\
    \includegraphics[width=5.5cm]{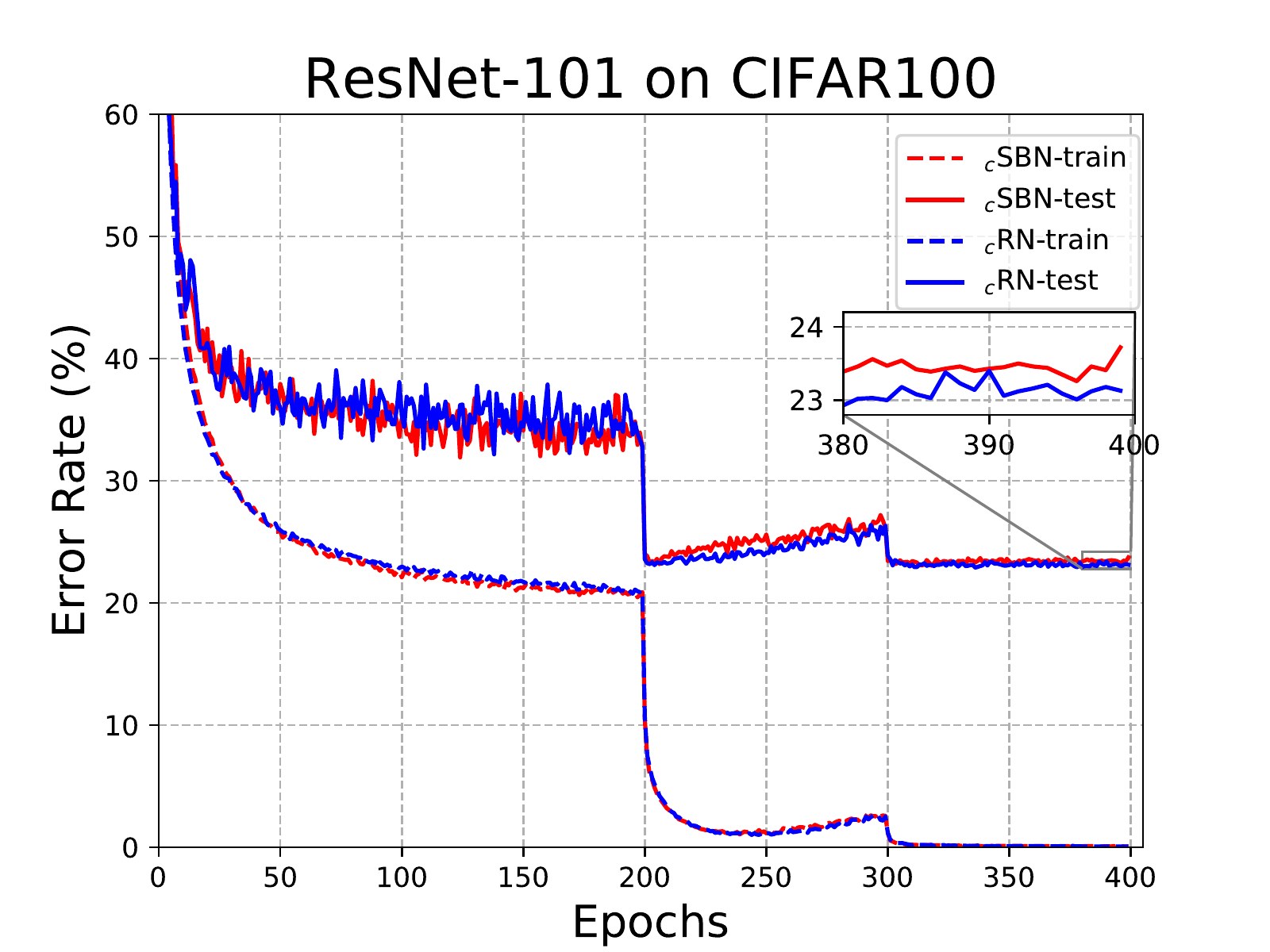}
    \includegraphics[width=5.5cm]{ResNetC101CIFAR100.pdf}
    \includegraphics[width=5.5cm]{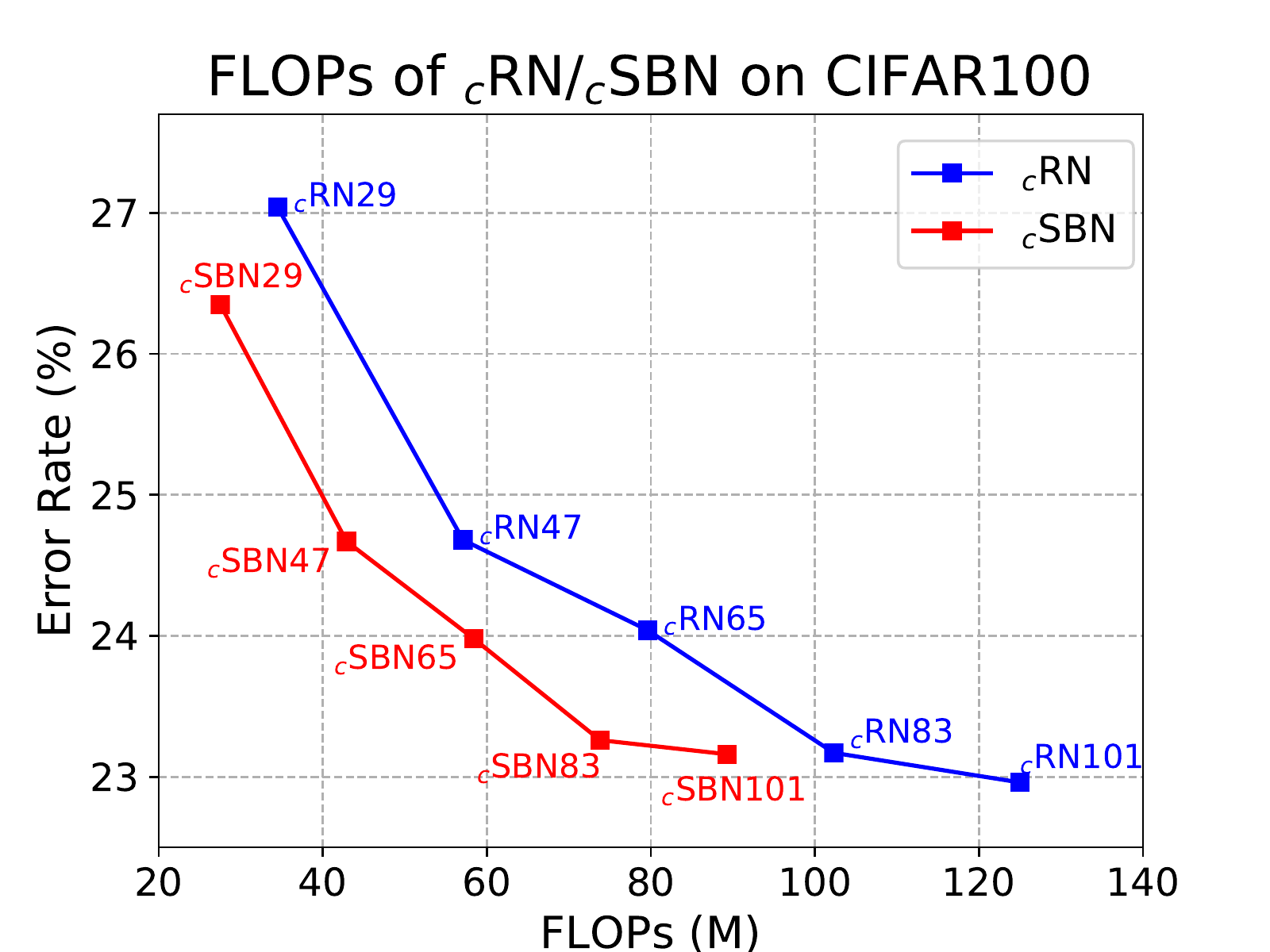}\\
\end{center}
\caption{
    The left two rows show the training and testing curves of our deepest networks ($110$ layers without channel bottleneck and $101$ layers with channel bottleneck) on the CIFAR10 and CIFAR100 datasets. Each case is randomly chosen from the $5$ individual runs. In each figure, we zoom-in on a small part for better visualization. The right column shows the relationship between the recognition error rate and the FLOPs, when networks with different depths are evaluated.
}
\label{Fig:CurvesCIFAR}
\end{figure*}

Results are summarized in Table~\ref{Tab:ResultsCIFAR}. Spatial bottleneck reduces the FLOPs of each network, but still leads to comparable classification accuracies\footnote{Out of $10$ comparisons ($5$ regular ResNets and $5$ channel-bottlenecked ResNets), spatial bottleneck works slightly better in $6$ case, and slightly worse in the remaining $4$. In addition, we verify that in each single case ($5$ individual runs for both the regular residual network and the spatial bottleneck network), we do not observe statistical significance ({\em i.e.}, ${p}\geqslant{0.05}$ using the student's $t$-test).}, and thus is more efficient when model complexity is taken into consideration (see the right column of Figure~\ref{Fig:CurvesCIFAR}). This implies that redundancy exists in the spatial domain, and reducing spatial redundancy is complementary to that in the channel domain (spatial bottleneck collaborates well with channel bottleneck~\cite{he2016deep}).

We plot the learning curves of different networks in Figure~\ref{Fig:CurvesCIFAR}. Basically, the curves before and after adding spatial bottleneck are very close to each other. In most scenarios, spatial bottleneck networks have higher training errors, which indicates that sparse sampling in the spatial domain harms the ability to fit training data. This is party caused by the small image sizes of the CIFAR datasets. In the ILSVRC2012 dataset (images are of higher resolutions), eliminating spatial redundancy is more effective, and so spatial bottleneck can achieve accuracy gain at the same time of acceleration.

\subsection{The Density of Spatial Sampling}
\label{ExperimentsCIFAR:Density}

\renewcommand{\colwidthA}{1.7cm}
\renewcommand{\colwidthB}{1.9cm}
\renewcommand{\colwidthC}{1.1cm}
\begin{table}[!btp]
\centering{
\setlength{\tabcolsep}{0.08cm}
\begin{tabular}{|l||R{\colwidthB}|R{\colwidthB}|R{\colwidthB}||R{\colwidthB}|}
\hline
$d$   & ${}_\mathrm{r}\mathrm{SBN}_{1/4}$ &  ${}_\mathrm{r}\mathrm{SBN}_{2/4}$ & ${}_\mathrm{r}\mathrm{SBN}_{3/4}$ &        ${}_\mathrm{r}\mathrm{RN}$ \\
\hline\hline
$20$  &                    $31.06\pm0.42$ &                    $30.57\pm0.23$ &                    $30.51\pm0.33$ &                    $30.10\pm0.39$ \\
\hline
$44$  &                    $28.39\pm0.20$ &                    $27.35\pm0.07$ &                    $27.65\pm0.19$ &                    $27.28\pm0.33$ \\
\hline
$62$  &                    $27.50\pm0.21$ &                    $26.27\pm0.28$ &                    $26.44\pm0.28$ &                    $26.28\pm0.32$ \\
\hline
$86$  &                    $26.49\pm0.34$ &                    $25.25\pm0.31$ &                    $25.85\pm0.13$ &                    $25.48\pm0.28$ \\
\hline
$110$ &                    $25.82\pm0.35$ &                    $25.13\pm0.01$ &                    $25.24\pm0.30$ &                    $25.01\pm0.32$ \\
\hline
\end{tabular}}
\caption{
    Classification error rate ($\%$) on the CIFAR100 dataset. Different network depths and sampling densities are considered. All the error rates reported are the average over $5$ individual runs.
}
\label{Tab:ResultsCIFARDensity}
\end{table}

\renewcommand{\colwidthA}{0.9cm}
\renewcommand{\colwidthB}{1.0cm}
\renewcommand{\colwidthC}{1.0cm}
\begin{table}[!btp]
\centering{
\setlength{\tabcolsep}{0.08cm}
\begin{tabular}{|l|l||R{\colwidthA}|R{\colwidthA}||R{\colwidthB}|R{\colwidthB}||R{\colwidthC}|R{\colwidthC}|}
\hline
\multicolumn{2}{|c||}{} &         \multicolumn{2}{c||}{CIFAR10} &                            \multicolumn{2}{c||}{CIFAR100} &                                \multicolumn{2}{c|}{FLOPs} \\
\hline
$d$   & $\alpha$ & ${}_\mathrm{r}\mathrm{RN}$ & ${}_\mathrm{r}\mathrm{SBN}$ &  ${}_\mathrm{r}\mathrm{RN}$ & ${}_\mathrm{r}\mathrm{SBN}$ &  ${}_\mathrm{r}\mathrm{RN}$ & ${}_\mathrm{r}\mathrm{SBN}$ \\
\hline\hline
$110$ & $84$     &                    $ 4.27$ &            $\mathbf{ 4.25}$ &            $\mathbf{20.21}$ &                     $20.64$ &           $ 1.15\mathrm{G}$ &   $\mathbf{ 0.60}\mathrm{G}$ \\
\hline
$110$ & $270$    &                    $ 3.73$ &            $\mathbf{ 3.53}$ &            $\mathbf{18.25}$ &                     $18.36$ &           $ 4.96\mathrm{G}$ &   $\mathbf{ 2.60}\mathrm{G}$ \\
\hline\hline
$164$ & $84$     &                    $ 3.96$ &            $\mathbf{ 3.80}$ &            $\mathbf{18.32}$ &                     $18.50$ &           $ 0.75\mathrm{G}$ &   $\mathbf{ 0.56}\mathrm{G}$ \\
\hline
$164$ & $270$    &           $\mathbf{ 3.48}$ &                     $ 3.67$ &            $\mathbf{17.01}$ &                     $17.20$ &           $ 2.92\mathrm{G}$ &   $\mathbf{ 2.16}\mathrm{G}$ \\
\hline
\end{tabular}}
\caption{
    Classification error rate ($\%$) and FLOPs of different pyramidal networks. Each architecture has almost the same FLOPs for CIFAR10 and CIFAR100.
}
\label{Tab:ResultsCIFARPyramid}
\end{table}

We perform several ablation studies to analysis the behaviors of our approach. Without loss of generality, we use the deep residual networks without channel bottlenecks and with different numbers of layers in these experiments.

We first verify the existence of spatial redundancy, which is our motivation and the reason that spatial bottleneck can reduce the sampling density without harming classification accuracies. Recall that spatial bottleneck ${}_\mathrm{r}\mathrm{SB}_{\left(a,b;K\right)}^{3\times3}$ partitions all $\left(w,h\right)$'s into $K$ subgroups, and samples on the one satisfying ${\left(w,h\right)}\equiv{\left(a,b\right)}\,{\left(\mathrm{mod}\,K\right)}$, thus preserving $1/K^2$ spatial information. We can increase the sampling density by considering more than one of the $K^2$ subgroups, {\em e.g.}, in ${K}={2}$, we used ${}_\mathrm{r}\mathrm{SB}_{\left(0,0;2\right)\vee\left(1,1;2\right)}^{3\times3}$ in the experiments of ${}_\mathrm{r}\mathrm{SBN}$. Continuing increasing the sampling density leads to $_\mathrm{r}\mathrm{SB}_{\left(0,0;2\right)\vee\left(0,1;2\right)\vee\left(1,1;2\right)}^{3\times3}$. For simplicity, we denote these variants by $_\mathrm{r}\mathrm{SBN}_{1/4}$, $_\mathrm{r}\mathrm{SBN}_{2/4}$ and $_\mathrm{r}\mathrm{SBN}_{3/4}$, respectively, in which the subscript indicates the number of sampled subgroups. $_\mathrm{r}\mathrm{SBN}_{4/4}$ is equivalent to the network with regular convolutions, therefore is not evaluated.

Results on networks of five different depths are summarized in Table~\ref{Tab:ResultsCIFARDensity}. We can observe that recognition accuracy {\em saturates} as the sampling density increases. In particular, when the fraction of sampled subgroups increases from $1/4$ to $2/4$, all the evaluated networks become more accurate significantly (${p}<{0.05}$). However, when the fraction continues to increase, the classification accuracies are not further improved. Similar phenomena are observed when channel bottleneck is present. Therefore, suggesting that sampling half of the spatial positions is able to preserve sufficient information. We inherit this strategy ($2/4$) in the ILSVRC2012 experiments.

\subsection{The Down-Sampling Strategy}
\label{ExperimentsCIFAR:DownSampling}

In the second ablation study, we investigate two approaches for down-sampling the input feature map. The first one is what we used before (using a stride-$K$ convolution), and the second one replaces it with a stride-$K$ average-pooling layer and a regular (stride-$1$) convolutional layer. The second version has a smaller number of parameters. The deconvolutional layer remains unchanged.

In experiments, the modified version reports unsatisfying classification performance. On the $110$-layer residual network, the original ${}_\mathrm{r}\mathrm{SBN}_{1/4}$ and ${}_\mathrm{r}\mathrm{SBN}_{2/4}$ achieve $25.82\pm0.35\%$ and $25.13\pm0.01\%$ error rates on CIFAR100, but the new version reports $26.28\pm0.38\%$ and $25.82\pm0.35\%$, respectively. The main accuracy drop is caused by the information loss brought by average-pooling. In comparison, a stride-$2$ convolution has the ability to preserve more spatial information although the down-sampled feature map has the same spatial resolution.

\subsection{Adjusting to Various Spatial Resolutions}
\label{ExperimentsCIFAR:PyramidNet}

The deep pyramidal residual networks~\cite{han2017deep} provided an effective configuration to increase the spatial resolution gradually within each section ({\em i.e.}, the same number of channels). We show that the spatial bottleneck block can adjust to PyramidalNet's. We consider the {\em add} version of the PyramidalNet (in which the spatial resolution goes up linearly with the depth). The network depth is set to be $110$ (without channel bottlenecks) and $164$ (with channel bottlenecks), and the $\alpha$ value (controlling the increasing rate of spatial resolution) is $84$ and $270$.

We train these networks for a total of $300$ epochs. The initial learning rate is set to be $0.1$ for CIFAR10 and CIFAR100, and is decayed by a factor of $0.1$ at $150$ and $225$ epochs, respectively. The other configurations remain the same as the above experiments. Results are summarized in Table~\ref{ExperimentsCIFAR:PyramidNet}. Similar to the situations on ResNet's, spatial bottleneck reduces the computational costs by $50\%$ and $27\%$ for the PyramidalNet's without and with channel bottlenecks, and the classification accuracies are comparable.

\section{ILSVRC2012 Experiments}
\label{ExperimentsILSVRC}

\renewcommand{\colwidthA}{1.9cm}
\renewcommand{\colwidthB}{1.0cm}
\begin{table*}[!btp]
\centering
\begin{tabular}{|l||R{\colwidthA}|R{\colwidthA}||R{\colwidthA}|R{\colwidthA}||R{\colwidthB}|R{\colwidthB}|}
\hline
      & \multicolumn{2}{c|}{Top-$1$ Error}                        & \multicolumn{2}{c|}{Top-$5$ Error}                        & \multicolumn{2}{c|}{FLOPs}                                \\
\hline
depth &  ${}_\mathrm{r}\mathrm{RN}$ & ${}_\mathrm{r}\mathrm{SBN}$ &  ${}_\mathrm{r}\mathrm{RN}$ & ${}_\mathrm{r}\mathrm{SBN}$ &  ${}_\mathrm{r}\mathrm{RN}$ & ${}_\mathrm{r}\mathrm{SBN}$ \\
\hline\hline
$18$  &              $30.70\pm0.07$ &     $\mathbf{30.20}\pm0.06$ &              $10.86\pm0.05$ &     $\mathbf{10.78}\pm0.04$ &            $ 1.6\mathrm{G}$ &   $\mathbf{ 1.0}\mathrm{G}$ \\
\hline
$34$  &              $26.82\pm0.06$ &     $\mathbf{26.48}\pm0.03$ &              $ 8.81\pm0.04$ &     $\mathbf{ 8.65}\pm0.02$ &            $ 3.5\mathrm{G}$ &   $\mathbf{ 2.0}\mathrm{G}$ \\
\hline
$50$  &              $24.52\pm0.07$ &     $\mathbf{23.96}\pm0.04$ &              $ 7.35\pm0.05$ &     $\mathbf{ 7.23}\pm0.03$ &            $ 3.7\mathrm{G}$ &   $\mathbf{ 2.9}\mathrm{G}$ \\
\hline
$101$ &              $23.16\pm0.19$ &     $\mathbf{22.69}\pm0.05$ &              $ 6.77\pm0.10$ &     $\mathbf{ 6.52}\pm0.03$ &            $7.4\mathrm{G}$ &    $\mathbf{ 5.7}\mathrm{G}$ \\
\hline
\end{tabular}
\caption{
    Classification error rate ($\%$) and FLOPs of different networks. The last $10$ snapshots (one at each epoch) are used for testing, and the averaged numbers as well as stand deviations are reported.
}
\label{Tab:ResultsILSVRC2012}
\end{table*}

\begin{figure*}[t]
\begin{center}
    \includegraphics[width=5.5cm]{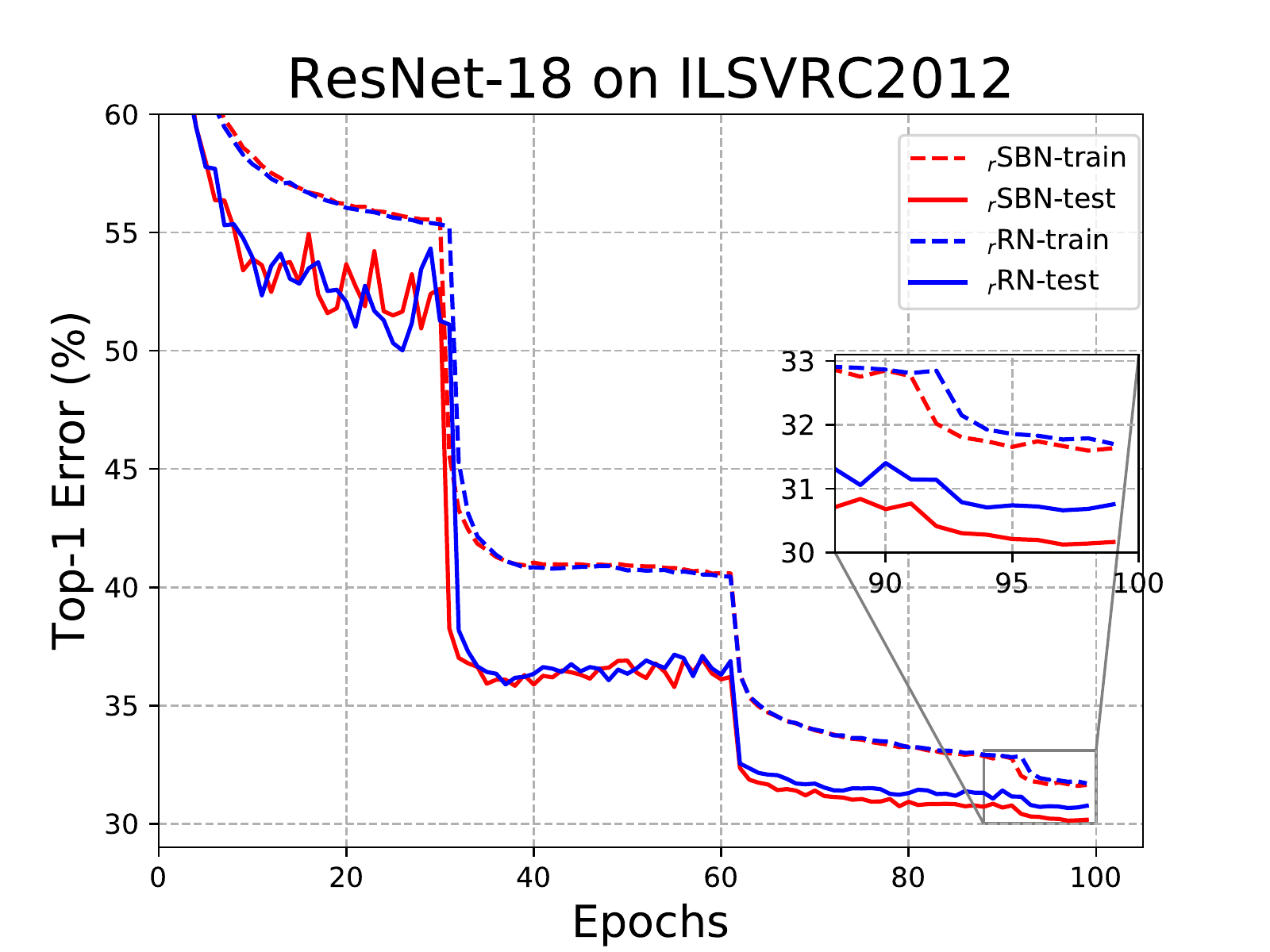}
    \includegraphics[width=5.5cm]{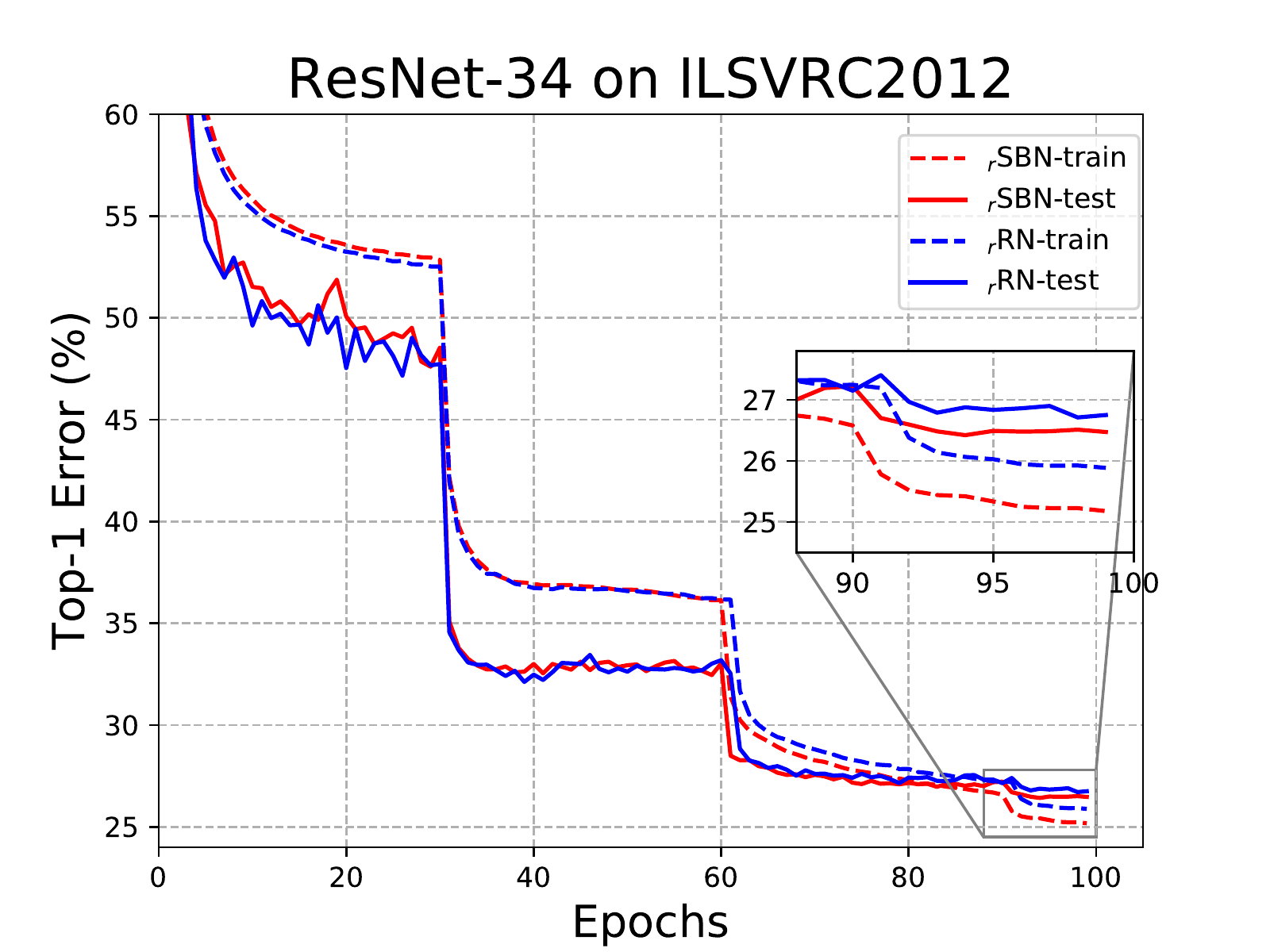}
    \includegraphics[width=5.5cm]{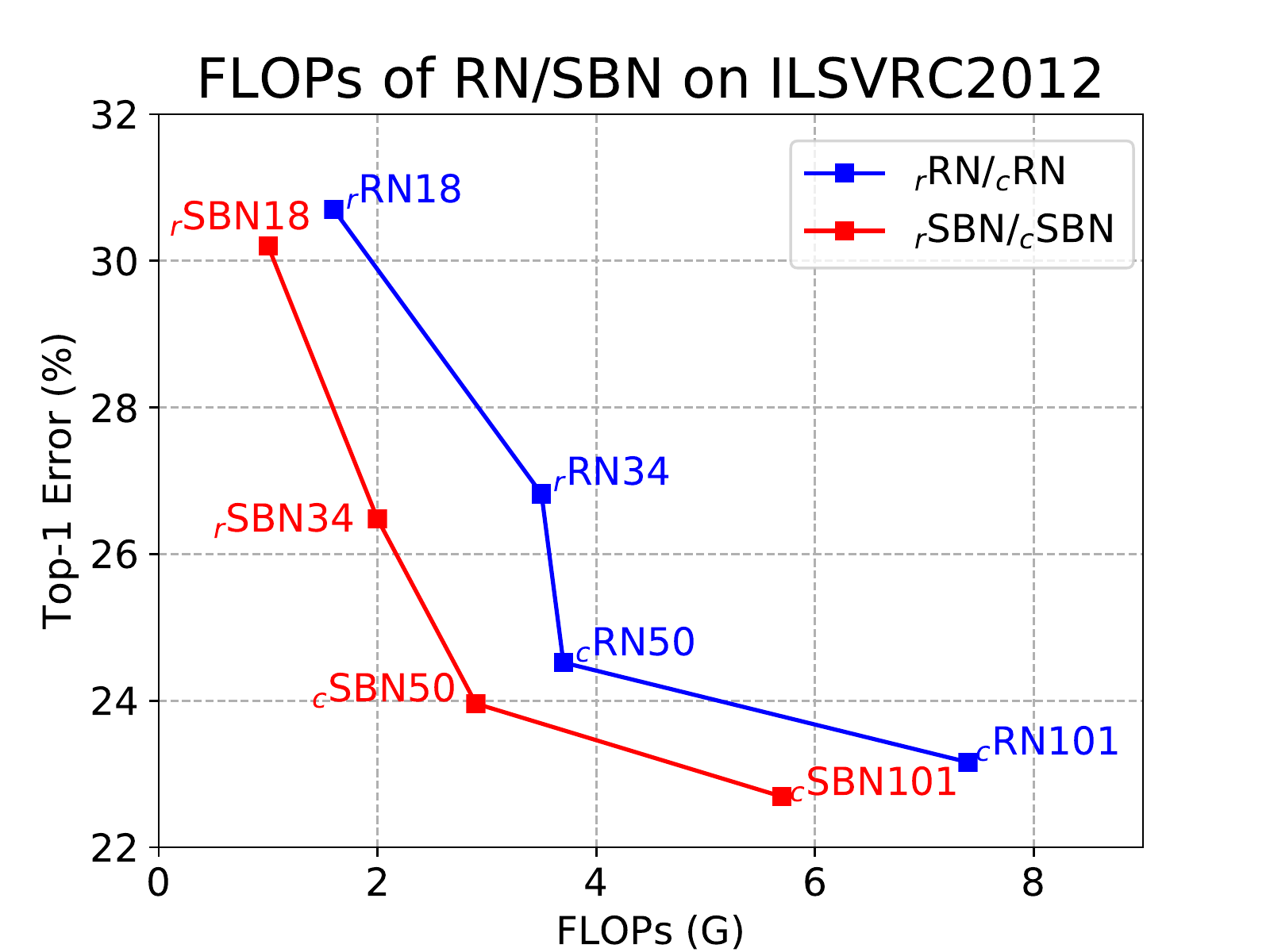}\\
    \includegraphics[width=5.5cm]{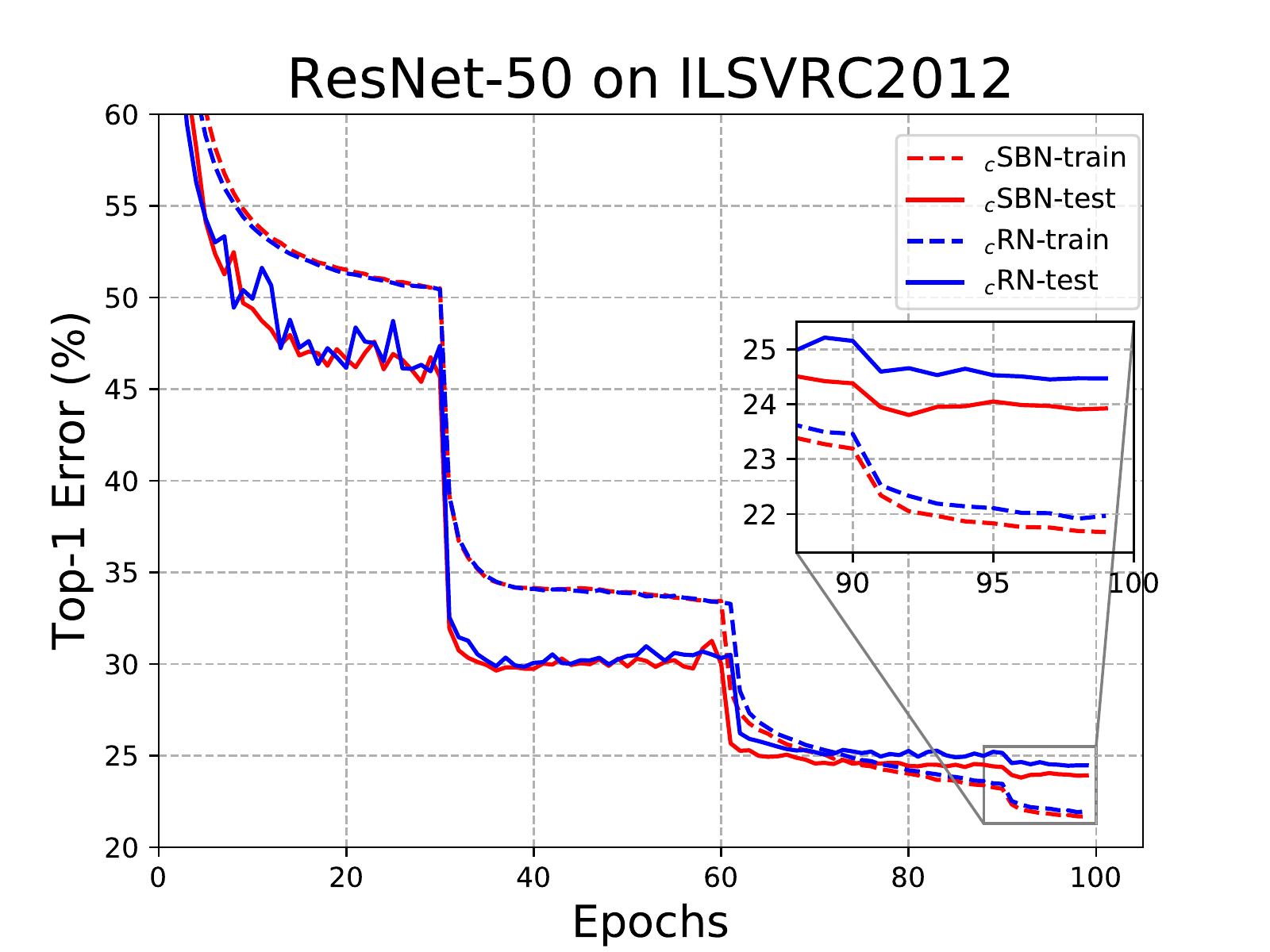}
    \includegraphics[width=5.5cm]{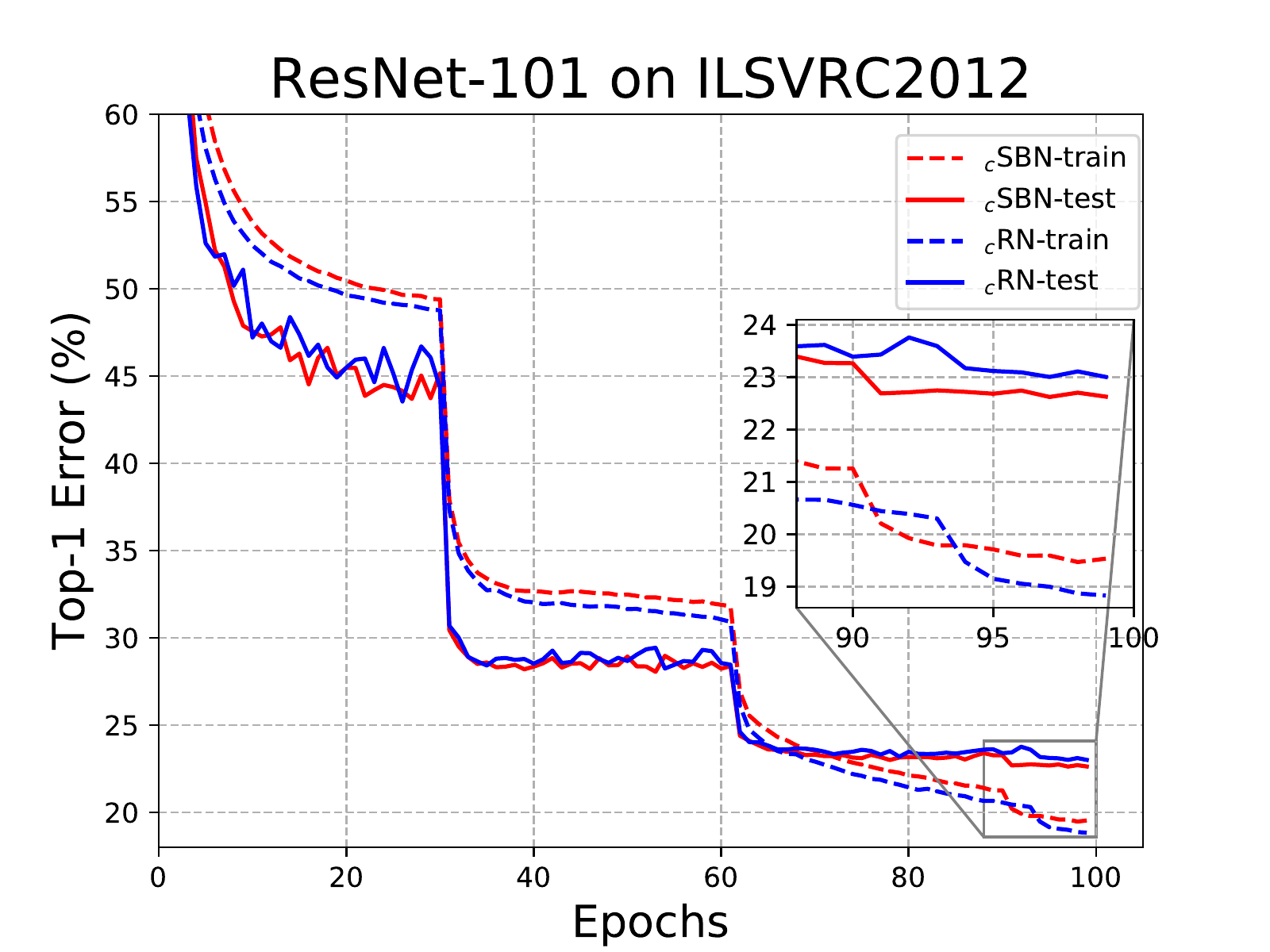}
    \includegraphics[width=5.5cm]{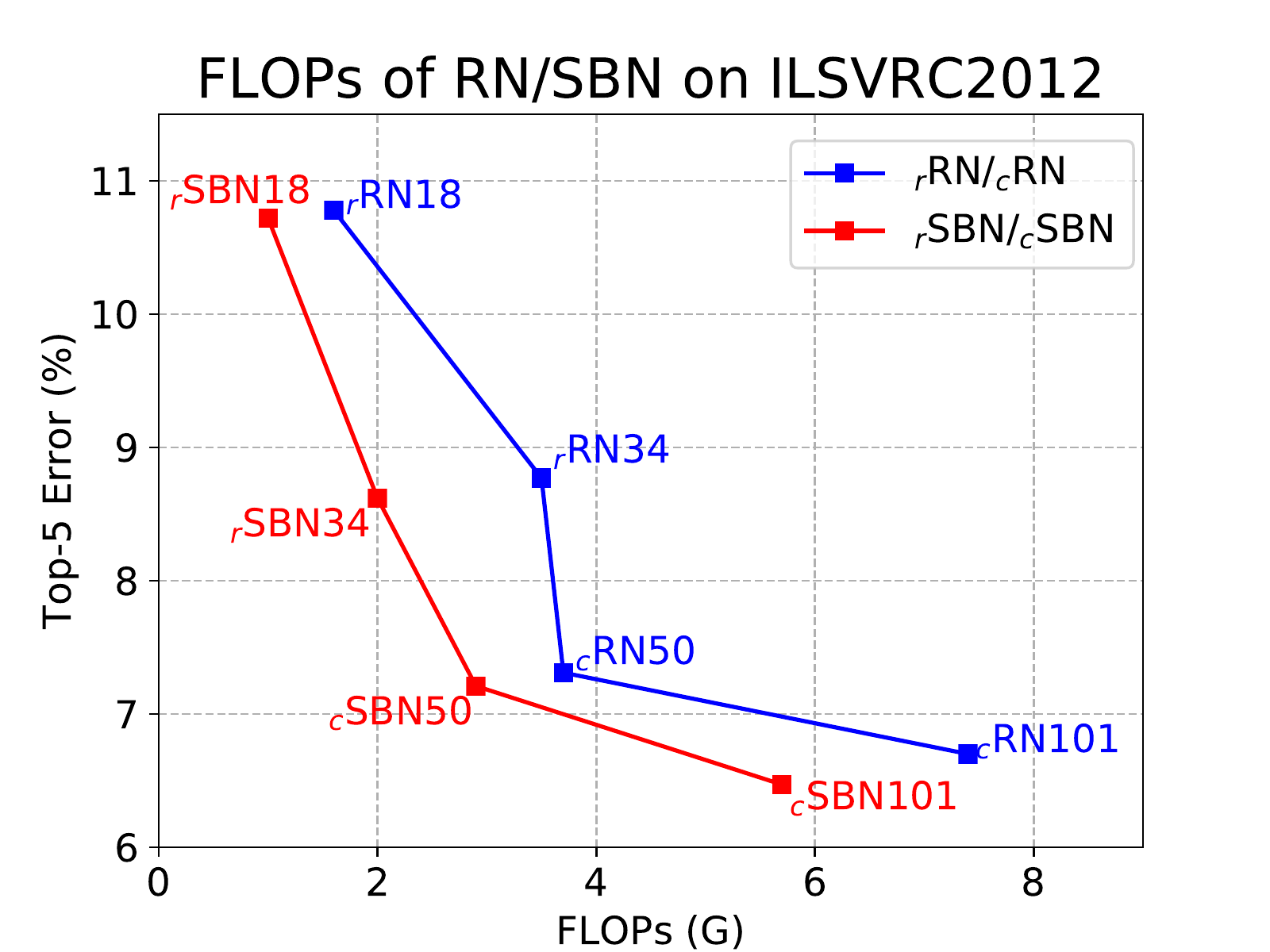}\\
\end{center}
\caption{
    The left two columns show the training and testing curves on ILSVRC2012 of deep residual networks with different numbers of layers. The top row shows two shallower networks without channel bottlenecks, and the bottom row two deeper ones with channel bottlenecks. For each case, we zoom-in on a small part for better visualization. The right column shows the relationship between the top-$1$ and top-$5$ recognition accuracies and the FLOPs.
}
\label{Fig:ILSVRC2012Curves}
\end{figure*}

\subsection{Settings}
\label{ExperimentsILSVRC:Settings}

The ILSVRC2012 dataset~\cite{russakovsky2015imagenet} is a subset of the ImageNet database~\cite{deng2009imagenet}. It contains $1\rm{,}000$ categories located at different levels of the WordNet hierarchy. The training set we use has $\sim1.3\mathrm{M}$ images roughly evenly distributed over all classes, and the testing set $50\mathrm{K}$ images, or exactly $50$ for each class.

We use the deep residual networks~\cite{he2016deep} with different layers, namely $18$, $34$, $50$ and $101$ layers. The former two are equipped with regular residual blocks, and the latter two channel bottleneck blocks. We replace each block, either with or without channel bottleneck, with the corresponding spatial bottleneck module. The fraction of saved computations is still $50\%$ for a regular residual block, and $27\%$ for a channel bottleneck block, respectively.

We train all these networks from scratch. Stochastic Gradient Descent (SGD) with a Nesterov momentum of $0.9$ is used. Each network is trained for $100$ epochs. The initial learning rate is set to be $0.1$, and divided by $10$ after $30$, $60$ and $90$ epochs. The size of each mini-batch is $256$, and the weight decay is $0.0001$. In the training stage, various data augmentation techniques are applied, including rescaling and cropping the image, randomly mirroring the image, changing its aspect ratio and performing pixel jittering. In the testing stage, a single crop of $224\times224$ is performed at the center of each image.

\subsection{Results}
\label{ExperimentsILSVRC:Results}

Results are summarized in Table~\ref{Tab:ResultsILSVRC2012}. Following the conventions, we report both top-$1$ and top-$5$ error rates for each model. We report the average and standard deviation of the testing accuracies of the final $10$ snapshots.

The spatial bottleneck block, after replacing each original residual block, consistently improves the classification performance. In particular, in the $34$-layer architecture, the top-$1$ and top-$5$ accuracies are boosted by $0.34\%$ and $0.16\%$, and the corresponding numbers are $0.47\%$ and $0.25\%$ in the $101$-layer architecture. The accuracy gain becomes more significant with the increased number of layers, {\em e.g.}, the relative top-$1$ and top-$5$ error rate drops are $1.63\%$ and $0.74\%$ on ResNet-18, and $2.03\%$ and $3.69\%$ on ResNet-101, respectively. These results verify the effectiveness of our building block in different network depths. Note that spatial bottleneck achieves these gains while reducing the computational costs of the baseline networks by either $50\%$ (without channel bottlenecks) or $27\%$ (with channel bottlenecks).

We plot the curves of top-$1$ and top-$5$ training/testing errors produced by the $34$-layer and $101$-layer deep residual networks in Figure~\ref{Fig:ILSVRC2012Curves}. Different from the curves on CIFAR100 (shown in Figure~\ref{Fig:CurvesCIFAR}), spatial bottleneck produces consistent gain in testing accuracy when the learning rate is sufficiently small (after the second decay at the $60$-th epoch). In addition, spatial bottleneck also achieves higher training accuracies except for the ResNet-101,  implying that a sparser spatial sampling does not harm the ability to fit training data.

Lastly, we plot the relationship between the testing accuracy and FLOPs in Figure~\ref{Fig:ILSVRC2012Curves}. Clearly, spatial bottleneck networks achieve better visual recognition performance with lower computational costs. This property makes our work stand out from some previous approaches~\cite{liu2015sparse}\cite{he2017channel} which accelerated the networks at the price of lower accuracies.

\section{Conclusions}
\label{Conclusions}

This paper presents spatial bottleneck, a simple and efficient approach to reduce the computational costs of a single convolutional layer or the combination of two convolutional layers. Spatial bottleneck works by temporarily reducing the spatial resolution using a stride-$2$ convolution and then restoring it using a stride-$2$ deconvolution. This is equivalently a regular way of sparsifying feature sampling in the spatial domain, and it is independent and complementary to such operations in the channel domain. We empirically verify that spatial bottleneck achieves comparable accuracies in CIFAR image classification and even higher accuracies in ImageNet image classification. Most importantly, the time costs in both training and testing are reduced significantly.

In this work, we only investigated the simplest case (${K}={2}$), in which the chessboard sampling strategy was verified effective. By increasing $K$, we can partition the spatial positions into more subgroups and thus allow a larger number of combinations of these subgroups. We believe that using a larger $K$ can improve the flexibility and recognition ability of our approach, and will investigate this topic in the future.

{\small
\bibliographystyle{ieee}
\bibliography{egbib}
}

\end{document}